# Dynamic LLM Routing and Selection based on User Preferences: Balancing Performance, Cost, and Ethics

Deepak Babu Piskala
Freshworks

Vijay Raajaa
Freshworks

Sachin Mishra
Freshworks

Bruno Bozza
Freshworks

**ABSTRACT**

With the widespread deployment of large language models (LLMs) such as GPT-4 [12], BART [9], and LLaMA [5], the need for a system that can intelligently select the most suitable model for specific tasks—while balancing cost, latency, accuracy, and ethical considerations—has become increasingly important. Recognizing that not all tasks necessitate models with over 100+ billion parameters, we introduce OptiRoute, an advanced model routing engine designed to dynamically select and route tasks to the optimal LLM based on detailed user-defined requirements. OptiRoute captures both functional (e.g., accuracy, speed, cost) and non-functional (e.g., helpfulness, harmlessness, honesty) criteria, leveraging lightweight task analysis and complexity estimation to efficiently match tasks with the best-fit models from a diverse array of LLMs. By employing a hybrid approach combining k-nearest neighbors (kNN) search and hierarchical filtering, OptiRoute optimizes for user priorities while minimizing computational overhead. This makes it ideal for real-time applications in cloud-based ML platforms, personalized AI services, and regulated industries. [4]

**General Terms**

LLM Optimization, Benchmarks, Evaluation, Routing, Complexity-estimation, Feedback, Domain Adaptation

**Keywords**

GPT4, Llama, Helpfulness, Honesty, Harmlessness, Latency, Accuracy, Cost, kNN, Optiroute, , Domain, Model Merging, Re-ranking, Fallback, Steerability, Instruction-following Ability, MLaaS, Healthcare, Finance, Legal, Hallucinations, Grounding, FLAN, BERT, BART

## 1. INTRODUCTION

The rapid advancement of large language models (LLMs) such as GPT-4 [12], BART [9], and LLaMA [5] has significantly transformed the field of natural language processing (NLP), enabling sophisticated applications across various sectors including healthcare, finance, legal services, and customer support. These models, with their vast number of parameters and deep learning architectures, have demonstrated state-of-the-art performance in tasks ranging from text generation and translation to sentiment analysis and complex multi-turn dialogues. However, their deployment in real-world applications presents substantial challenges that limit their accessibility and efficiency, particularly for organizations with constrained resources.

With Huggingface [14] hosting over 486,000 foundational and fine-tuned models, and more than 1,000 new models added daily, the challenge of discovering the right model for a specific task has become increasingly complex. This vast and rapidly growing repository caters to a wide range of domains, accuracy levels, architectures, and tasks. However, only a small percentage of trending or popular models are effectively utilized, leaving the vast majority of models underexplored and underutilized. This underscores the critical need for advanced systems that can intelligently navigate this immense landscape, ensuring that the most suitable models are identified and deployed based on user-specific criteria and task requirements

One of the primary challenges associated with deploying LLMs is the high computational cost and resource demand they impose. Running models with hundreds of billions of parameters requires significant processing power, leading to increased latency and operational expenses, especially in cloud-based environments where compute resources are billed by usage. This is particularly problematic for applications requiring real-time or near-real-time responses, such as interactive chatbots, automated trading systems, or autonomous vehicles, where even slight delays can be detrimental. Additionally, the one-size-fits-all approach commonly employed in deploying LLMs—where a single, often overpowered model is used for all tasks—fails to account for the varying complexities of different tasks. Not all tasks necessitate the full power of models with over 100 billion parameters; simpler tasks could be effectively handled by smaller, more cost-efficient models, thereby reducing unnecessary resource consumption.

Beyond technical and cost considerations, the ethical deployment of AI has emerged as a critical concern. As LLMs are increasingly integrated into applications that directly interact with humans, ensuring that these models behave in ways that are honest, harmless, and helpful [1] is paramount. The lack of mechanisms to incorporate ethical considerations into model selection can lead to significant risks, including the propagation of biased or harmful content, erosion of user trust, and potential regulatory repercussions. Current systems often overlook these ethical dimensions, focusing primarily on performance metrics like accuracy or speed,without addressing how the outputs align with broader societal values and norms.

To address these multifaceted challenges, we introduce **OptiRoute**, a novel model routing engine designed to optimize the deployment of LLMs by dynamically selecting and routing tasks to the most appropriate model based on detailed user-defined requirements. OptiRoute captures both functional requirements—such as accuracy, speed, and cost—and non-functional requirements, including ethical considerations like helpfulness, harmlessness, and honesty [1]. By tailoring the model selection process to the specific needs of each task and user, OptiRoute ensures that the chosen model not only meets performance expectations but also aligns with ethical standards.

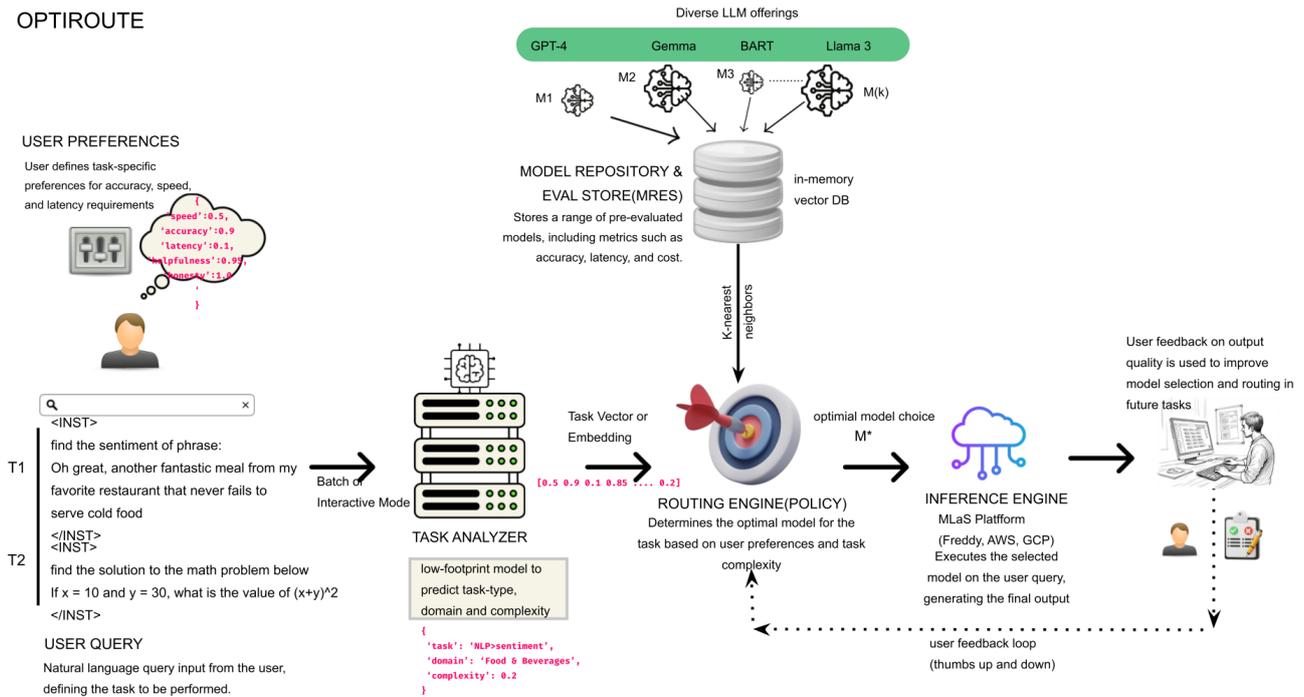

Fig. 1: OptiRoute Architecture: (i) Captures user-defined functional and non-functional requirements to tailor the system's response. (ii) Analyzes task complexity through a lightweight Task Analyzer, generating a task vector that is used for model selection. (iii) Employs a kNN-based Routing Engine to match tasks with the optimal LLM from a diverse model repository eval (MRES), followed by real-time inference execution and a feedback loop to refine future selections. The entire process ensures efficient, ethical, and cost-effective LLM deployment tailored to user needs

**OptiRoute** operates through a hybrid approach that integrates lightweight task analysis and complexity estimation with a k-nearest neighbors (kNN) search mechanism. Initially, the system analyzes the incoming task to determine its type, domain, and complexity, generating a task vector that encapsulates these characteristics. This task vector is then used in conjunction with user-defined preferences to search the Model Repository and Evaluation Store (MRES), an in-memory vector database that houses a diverse array of pre-evaluated models, ranging from low-cost, open-source alternatives to high-performance, proprietary LLMs. The kNN-based Routing Engine leverages this database to identify the optimal model for the task, balancing the trade-offs between cost, performance, and ethical considerations.

This dynamic routing and selection process offers several key advantages. First, by selecting models based on the specific complexity of the task, OptiRoute minimizes unnecessary computational overhead, reducing latency and cost. Second, the system's ability to integrate user preferences and ethical criteria into the model selection process ensures that the deployed AI behaves in a manner consistent with user expectations and societal norms. Third, the inclusion of a feedback loop, where user feedback on the output quality is used to refine future model selections, enables continuous optimization and adaptation of the system, further enhancing its reliability and effectiveness. Fig 1 shows the architecture of the proposed system. In the next few sections, we deep-dive into the individual components and end-to-end orchestration of the system. We also cover use-cases and potential extensions of this work both from a research and applied perspective. Finally, we show the potential benefits of deploying these systems at an enterprise level ML platform at Freshworks called Freddy ML.

## 2. APPLICATIONS OR USE-CASES

—**Cloud-Based Machine Learning Platforms (MLaaS)**: OptiRoute can be integrated into MLaaS platforms like AWS, Google Cloud, or Azure to optimize the selection and deployment of LLMs based on user-specific criteria such as cost, latency, accuracy, and ethical considerations. This ensures efficient resource utilization, reduces operational costs, and enhances the performance and reliability of AI services in the cloud, particularly for applications like chatbots that require balancing high accuracy with budget constraints.

—**Personalized AI Services**: In personalized AI services such as recommendation engines, virtual assistants, and targeted mar-keting, OptiRoute can tailor model selection to align with individual user preferences. By dynamically routing tasks to models that meet specific needs—whether prioritizing speed, accuracy, or privacy— OptiRoute enhances user satisfaction, engagement, and delivers a more personalized and effective interaction in consumer-facing applications. [16]

—**Regulated Industries (Healthcare [11], Finance [17] [18], Legal [10])**: OptiRoute is ideal for regulated industries like healthcare, finance, and legal services, where accuracy, security, and ethical compliance are paramount. It can route tasks such as medical diagnostics or financial trading to models optimized for high accuracy and regulatory compliance, while also ensuring ethical AI behavior, thereby improving the reliability of critical applications and mitigating risks associated with unethical practices.

—**Data Annotation and Labeling for AI Training**: In AI training processes that require large volumes of labeled data, OptiRoute can optimize data annotation platforms by routing tasks to models best suited for specific types of data (e.g., text, images, video) and label accuracy requirements. Simple annotation tasks might be handled by fast, cost-effective models, while more complex or ambiguous cases can be routed to models with higher accuracy and deeper understanding, thereby enhancing the efficiency and accuracy of data labeling processes and reducing the time and cost needed to produce high-quality training data for AI models

## 3. SYSTEM DESIGN ARCHITECTURE

OptiRoute offers two distinct modes of operation—batch and interactive—each tailored to different user needs and operational contexts. In batch mode, users submit a collection of queries along with their preferred optimization criteria, such as cost, speed, or accuracy. To optimize efficiency, OptiRoute samples a small percentage (typically 2the queries to determine the most suitable large language model (LLM) that can effectively handle the entire batch. This mode is particularly effective when dealing with large volumes of relatively homogeneous queries, as it minimizes computational overhead by avoiding the need for individual query assessments, making it ideal for offline processing or scheduled tasks.

On the other hand, interactive mode is designed for realtime query assessment, where each query is individually analyzed and routed to the best-suited LLM based on the user's specified criteria. This mode is ideal for scenarios requiring immediate and precise responses, such as customer service bots or virtual assistants, where the system must dynamically adapt to each unique query. While interactive mode demands more computational resources, it provides maximum flexibility and ensures that each query receives a tailored response optimized for accuracy, latency, and other user-defined factors.

The choice between batch and interactive modes allows users to balance the need for efficiency with the requirement for real-time precision, depending on their specific application. Batch mode excels in processing large volumes of similar queries efficiently, while interactive mode offers superior adaptability and responsiveness for dynamic, userfacing applications. This flexibility ensures that OptiRoute can cater to a wide range of operational scenarios, from large-scale data processing to real-time AI interactions

### 3.1 User Preferences

Let us try to define "user" and "preferences" in detail. Preferences can be classified as being either explicit (i.e mentioned by user) or implicit (i.e unsaid but expected behavior from user). Explicit preferences include ability to provide scores from 0 (low) to 1(high) for functional requirements like accuracy, speed, cost and non-functional requirements like helpfulness, honesty and harmlessness. We use a lightweight task analyzer to determine task-type and query complexity, which can be considered as unsaid preferences by the user but automatically inferred from the query. Fig 2 shows the exhaustive list of preferences that can be specified or inferred from the user. The user in this context could assume multi-ple roles (i) end-user i.e one who submits a query to MLaaS cloud provider or (ii) Admin who configures this for a cohort of users or batch, who could be an AI engineer or MLE. We expect an average end-user to be not fully aware of the all knobs and sliders exposed in optiroute, hence from an UX perspective we also offer profiles which encapsulate complex combinations of settings to easily relatable user preferences. A few examples of such a profile include "cost-effective", "ethically-aligned", "latency-first", etc.

| **Explicit** | **Implicit** |
|---|---|
| Accuracy | Task-type |
| Latency | Complexity |
| Cost | Domain |
| Harmlessness | |
| Honesty | |
| Helpfulness | |
| Steerability | |
| Creativity | |

Table 1. : User Preferences

### 3.2 Task Analyzer

Task Analyzer is a low-footprint ML model (400M autoregressive encoder-decoder language model like FLAN-T5 [2]) that is instruction fine-tuned (IFT) to predict implicit preferences and characteristics of query at run-time for efficient routing based on user preferences. The output of the LLM is a structured json with fields that predict (i) task type (ii) domain and (iii) complexity and can be extended to add more tasks. The data for fine-tuning is collected through a mix of supervised and synthetic techniques like self-align and self-instruct. We use the query logs of a production MLaaS cloud provider to label a random sample of N queries through a combination of human annotations and semi-supervised learning (SSL). To further optimize the latency of the task analyzer model, we consider quantization techniques like reducing precision to 4-bit or 8-bit [19] that reduces memory and compute demands. Another optimization we consider owing to quadratic complexity of response time as a function of input token, we consider custom pruning logic in case of long queries. Since the query word count could vary widely from few 50 words to 10K+ words with large context blobs, we consider pruning the long query text to only consider the first n and last n words which usually contains the task description like 'find the sentiment of the passage' and random sample sentences or words from the middle portion of query. Fig 3 shows the json response for a sample query for a sentiment analysis task.

### 3.3 Model Registry Evaluation Store (MRES)

The Model Registry and Evaluation Store (MRES) is a critical component of the OptiRoute system, serving as the centralized repository where all available models are stored, evaluated, and accessed during the model selection process. The MRES is designed to maintain a comprehensive inventory of large language models (LLMs), including both proprietary and open-source models, each annotated with a variety of performance and ethical metrics. This repository not only stores the models themselves but also crucial metadata and evaluation results that enable OptiRoute to dynamically route tasks to the most appropriate model based on user-defined criteria and task complexity.

At its core, the MRES functions as an in-memory vector database, optimized for fast retrieval and efficient storage of model information. Each entry in the MRES represents an individual LLM, along with detailed metadata that describes the model's architecture, parameter count, performance across various benchmarks, and other key characteristics. The evaluation data stored in the MRES is comprehensive, covering a wide range of metrics that are crucial for making informed model selection decisions. These metrics include accuracy, inference time, cost per inference, ethical considerations (such as helpfulness, harmlessness, and honesty), security and privacy features, and reliability (e.g., uptime percentage).

Given that the metrics gathered during evaluation can vary significantly in scale (e.g., accuracy as a percentage, cost in dollars, inference time in milliseconds), a normalization process is employed to standardize these metrics. This normalization logic converts each metric into a standard range of 0 to 1, enabling relative comparisons across models. For instance, a model with the highest accuracy might receive a normalized accuracy score of 1, while a model with slower inference times might receive a lower score on the speed metric. This standardized format allows the Routing Engine to easily compare models on a like-for-like basis, ensuring that the most suitable model is selected according to the user's explicit and implicit preferences.

### 3.4 Routing Engine

The Routing Engine is a central component of the OptiRoute system, responsible for dynamically selecting the most suitable large language model (LLM) from the available catalog in the Model Registry and Evaluation Store (MRES). This selection process is driven by advanced techniques, including query embedding, approximate k-nearest neighbors (kNN) search, filtering, and scoring. These methods ensure that each task is matched with the optimal model based on user-defined requirements and the specific characteristics of the task.

When a user submits a query, the Task Analyzer processes the input to determine its type, domain, and complexity, resulting in the generation of a task vector, also referred to as a query embedding. This task vector is a numerical representation that encapsulates the essential features of the task, such as the required accuracy, complexity, domain specificity (e.g., legal, medical), and any ethical considerations like harmlessness or honesty. The task vector serves as the input for the Routing Engine, providing a compact and computationally efficient way to represent the task in the context of model selection. The Routing Engine then uses this vector to search the MRES, an in-memory vector database that stores pre-computed embeddings for all models in the catalog. Each model's embedding represents its capabilities across various metrics and task types, allowing the system to compare and select models effectively.

To identify the most suitable models, the Routing Engine employs an approximate k-nearest neighbors (kNN) search algorithm within the MRES. Approximate kNN is chosen for its balance between search accuracy and computational efficiency, making it ideal for real-time applications where quick decision-making is crucial. The kNN search works by comparing the task vector against the pre-computed model embeddings stored in the MRES, identifying the top k models whose embeddings are closest to the task vector. This proximity is determined based on the normalized metrics stored in the MRES, such as accuracy, speed, cost, and ethical considerations, ensuring that the selected models are the best candidates for handling the specific query.

After the top k models are identified through the kNN search, the Routing Engine applies additional filtering and scoring mechanisms to further refine the selection process. The initial filtering process ensures that only models relevant to the specific task type are considered. For example, if the task involves legal document processing, models not specialized in legal NLP tasks are filtered out, narrowing the candidates to those with demonstrated expertise in the relevant domain. Further filtering is applied based on domain specificity, ensuring that only models tagged in the MRES as particularly effective in the required domain are retained. Following this, the remaining models are scored based on their normalized metrics, with weights assigned according to the user's explicit preferences. This scoring system ensures that the model most aligned with the user's priorities is selected, balancing performance with resource efficiency.

In scenarios where no exact matching model is found—perhaps due to a highly specialized or novel task—the Routing Engine is designed to handle such cases through fallback mechanisms. The system may resort to using generalist models, which are versatile LLMs capable of handling a wide range of tasks with reasonable performance. Alternatively, the Routing Engine might expand the kNN search to include models that, while not exact matches, still exhibit a relatively high degree of alignment with the task vector. These fallback models provide the best available options under the circumstances, ensuring that the task can still be effectively processed. In cases where only fallback models are available, the system can notify the user, allowing them to adjust their preferences or provide additional input to refine the search. The system can also adapt by learning from these scenarios, improving future model selections as the MRES database evolves. The Routing Engine relies on a distance metric to evaluate the similarity between the task vector and the pre-computed model embeddings stored in the MRES. The most commonly used distance metric in this context is the cosine similarity, which measures the cosine of the angle between two non-zero vectors in a multi-dimensional space. Cosine similarity is particularly well-suited for comparing high-dimensional vectors like those used in task embeddings because it focuses on the orientation of the vectors rather than their magnitude, making it robust against variations in scale across different metrics. In the context of OptiRoute, the cosine similarity metric is used to calculate the distance between the task vector and each model's embedding. The model with the lowest cosine distance—indicating the smallest angle and therefore the greatest similarity between the task requirements and the model's capabilities—is selected as the optimal model. This approach ensures that the model chosen is the one most aligned with the specific characteristics and demands of the task, providing a precise and reliable method for model selection.

### 3.5 User Feedback and Inference Engine

The inference engine in **OptiRoute** is responsible for executing the selected model on the user's query and delivering the final output. This process is crucial as it represents the final step in the model selection and task routing workflow, ensuring that the chosen model

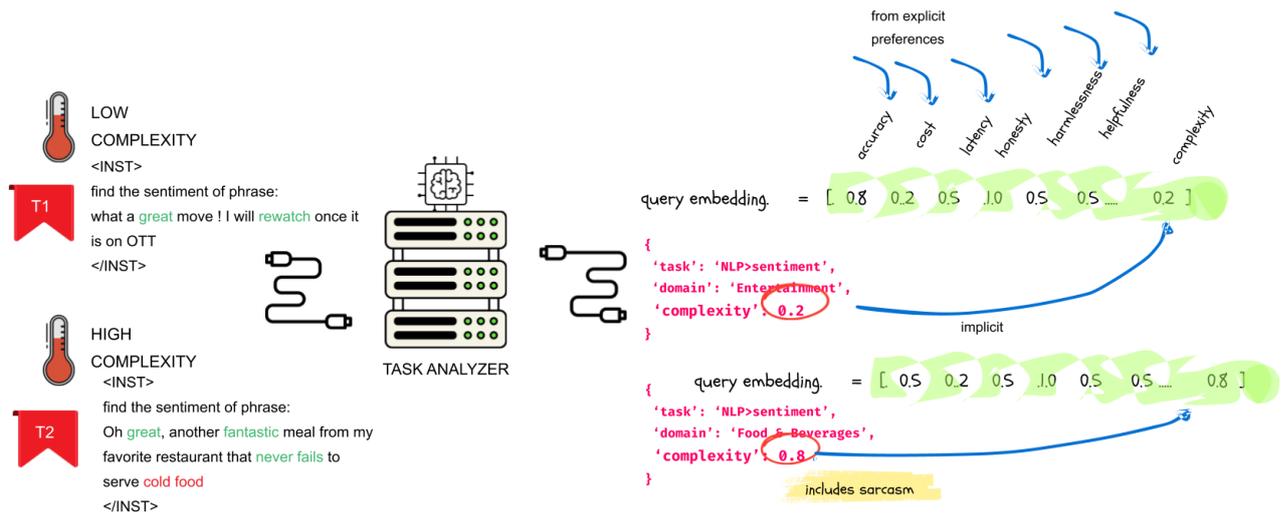

Fig. 2: This diagram illustrates the task complexity analysis and embedding generation process in the OptiRoute system. Two sentiment analysis tasks with varying complexities are analyzed by a Task Analyzer. Task T1, categorized as low complexity (0.2), involves a straightforward, positive sentiment, while Task T2, a high complexity task (0.8), contains sarcasm and more nuanced sentiment within the "Food Beverages" domain. The Task Analyzer generates query embeddings based on both explicit preferences (e.g., accuracy, cost, latency, honesty) and implicit preferences inferred from the task itself, helping to dynamically route the tasks to the most appropriate LLMs that align with the user's functional and non-functional requirements

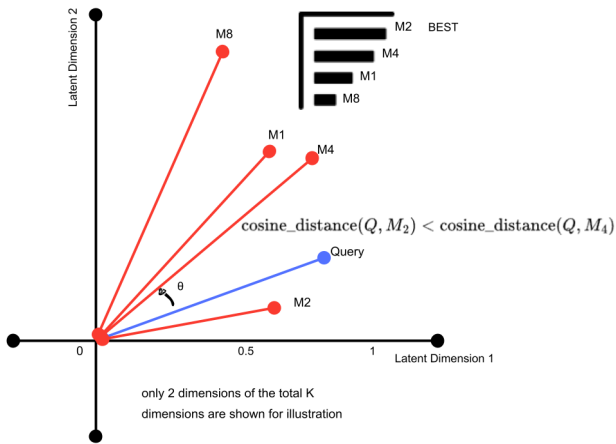

Fig. 3: This diagram illustrates the cosine distance-based model selection process in the OptiRoute system. The query vector (Q) is represented in a latent space along with multiple pre-evaluated models (M1, M2, M4, M8). The cosine distance between the query and each model's embedding is calculated, with the smaller angle indicating a closer match. In this example, model M2 has the smallest cosine distance to the query, making it the most suitable choice, as shown in the ranking bar chart. Only two dimensions of the full latent space are shown for illustration purposes

performs the task according to the specified user preferences. However, the efficacy of this system does not end with the inference; it is further enhanced by incorporating user feedback into the routing policy, enabling continuous improvement of the model selection process.

User feedback, typically captured through mechanisms such as thumbs up or thumbs down, plays a pivotal role in refining the routing policy. After the inference engine delivers the output, users are encouraged to provide feedback on the quality and relevance of the results. A thumbs-up indicates that the output met or exceeded the user's expectations, confirming that the selected model was appropriate for the task. Conversely, a thumbs-down signals that the output was suboptimal, suggesting that the chosen model may not have been the best fit for the query or that the model failed to align with the user's explicit or implicit preferences.

This feedback loop is instrumental in improving the routing policy over time. When the system receives positive feedback (thumbs up), it reinforces the current model selection strategy, making it more likely that similar queries in the future will follow the same routing path. On the other hand, negative feedback (thumbs down) triggers a review of the decision-making process. The system can analyze the characteristics of the task vector, the chosen model, and the resulting output to identify why the model underperformed. This analysis might reveal that the model's capabilities did not match the task's complexity, that the domain specificity was inadequate, or that user preferences such as ethical considerations were not fully met.

## 4. RELATED WORK

Several recent studies have explored leveraging multiple large language models (LLMs) for complex, multi-step tasks, most notably HuggingGPT [13]. HuggingGPT utilizes multiple models from the Hugging Face model repository to solve intricate tasks by invoking different specialized models for subtasks. For example, in the case of translating English audio to German, HuggingGPT first uses an automatic speech recognition (ASR) model to transcribe the speech, then a translation model to convert the text into German, and finally a text-to-speech (TTS) model to generate audio in German. While HuggingGPT focuses on decomposing tasks across different models, our approach in OptiRoute is to similarly select and route tasks to the most suitable LLM based on user preferences, but with an emphasis on balancing functional and non-functional requirements, such as cost, latency, accuracy, and ethical considerations, without necessarily solving new tasks.

In terms of inference efficiency, the literature has increasingly focused on reducing the computational overhead of LLMs during inference. Techniques like model quantization [8], which reduces the precision of the model weights to lower memory usage and speed up computations, have been widely adopted. These methods directly align Dynamic LLM Routing and Selection Based on User Preferences with our goal of minimizing computational costs in realtime model selection, making them integral to OptiRoute's efficiency optimizations.

Another related line of research includes parameter-efficient fine-tuning methods such as LoRA [7] and qLoRA [3], which reduce the number of trainable parameters during finetuning by applying low-rank adaptations to pre-trained models. These techniques enable faster fine-tuning with less computational cost, a strategy that complements OptiRoute's focus on selecting models based on resource constraints and task complexity.Approaches such as model averaging and model soups [15], where multiple fine-tuned models are combined by averaging their weights, have also gained traction in improving generalization performance across diverse tasks. This idea resonates with OptiRoute's objective to dynamically route tasks to the most appropriate models from a diverse repository of LLMs, which may involve averaging models to meet user-specified criteria.

Knowledge distillation [6] is another relevant area that deals with compressing large models into smaller, more efficient versions without significant loss in performance. By transferring knowledge from a large "teacher" model to a smaller "student" model, the distillation process enables efficient inference in resource-constrained environments. This concept parallels our goal of optimizing model selection to strike a balance between cost, accuracy, and latency, particularly for simpler tasks that do not require the full capacity of the largest LLMs. Finally, sparse mixture of experts (MoE) models propose another avenue for dynamic and efficient model routing. By activating only a subset of model experts for each task, MoEs can dramatically reduce the computation needed for inference while maintaining high accuracy. While OptiRoute doesn't directly implement sparse MoE models, the underlying principle of activating models selectively based on task complexity is akin to our model routing mechanism.

To the best of our knowledge, no prior work has explored the personalization of LLM routing based on both functional and non-functional requirements—such as accuracy, cost, latency, and ethical dimensions—while dynamically optimizing trade-offs across over 10 parameters, including helpfulness, steerability, and hallucination risk. OptiRoute stands out by introducing a hybrid task analysis and routing engine that integrates user-defined preferences to ensure efficient and personalized model selection, optimizing LLM deployment in real-world applications.

## 5. FUTURE DIRECTIONS

A promising future direction for OptiRoute is the development of a capability to merge models in cases where no existing model fully meets the user-specified criteria. When the system encounters a scenario where no single model perfectly aligns with the user's requirements—whether due to limitations in accuracy, speed, cost, or ethical considerations— OptiRoute could dynamically generate a new model by merging the weights of multiple individual models that each partially meet the criteria. This approach would involve selectively combining the strengths of different models, effectively creating a hybrid model tailored to the specific task at hand. For example, if one model excels in accuracy but is costly, and another is cost-efficient but less accurate, OptiRoute could merge their respective weights to achieve a balance that better aligns with the user's needs. This process, which could leverage techniques from model ensembling, transfer learning, and low-rank adaptation (LoRA), would enable the creation of novel models on-the-fly, expanding the system's flexibility and capability to deliver highly customized solutions even in complex, multi-faceted scenarios where pre-existing models fall short.

## 6. REFERENCES

[1] Amanda Askell, Yuntao Bai, Anna Chen, Dawn Drain, Deep Ganguli, Tom Henighan, Andy Jones, Nicholas Joseph, Ben Mann, Nova DasSarma, Nelson Elhage, Zac Hatfield-Dodds, Danny Hernandez, Jackson Kernion, Kamal Ndousse, Catherine Olsson, Dario Amodei, Tom Brown, Jack Clark, Sam McCandlish, Chris Olah, and Jared Kaplan. A general language assistant as a laboratory for alignment. *arXiv preprint arXiv:2112.00861*, December 2021.

[2] Hyung Won Chung, Le Hou, Shayne Longpre, Barret Zoph, Yi Tay, William Fedus, Yunxuan Li, Xuezhi Wang, Mostafa Dehghani, Siddhartha Brahma, Albert Webson, Shixiang Shane Gu, Zhuyun Dai, Mirac Suzgun, Xinyun Chen, Aakanksha Chowdhery, Alex Castro-Ros, Marie Pellat, Kevin Robinson, Dasha Valter, Sharan Narang, Gaurav Mishra, Adams Yu, Vincent Zhao, Yanping Huang, Andrew Dai, Hongkun Yu, Slav Petrov, Ed H. Chi, Jeff Dean, Jacob Devlin, Adam Roberts, Denny Zhou, Quoc V. Le, and Jason Wei. Scaling instruction-finetuned language models, 2022.

[3] Tim Dettmers, Artidoro Pagnoni, Ari Holtzman, and Luke Zettlemoyer. Qlora: Efficient finetuning of quantized llms. *arXiv preprint arXiv:2305.14314*, May 2023.

[4] Elias Frantar, Saleh Ashkboos, Torsten Hoefler, and Dan Alistarh. Gptq: Accurate post-training quantization for generative pre-trained transformers, 2023.

[5] Aaron Grattafiori, Abhimanyu Dubey, and Abhinav Jauhri et. al. The llama 3 herd of models, 2024.

[6] Geoffrey Hinton, Oriol Vinyals, and Jeff Dean. Distilling the knowledge in a neural network. *arXiv preprint arXiv:1503.02531*, March 2015.

[7] Edward J. Hu, Yelong Shen, Phillip Wallis, Zeyuan Allen-Zhu, Yuanzhi Li, Shean Wang, Lu Wang, and Weizhu Chen. Lora: Low-rank adaptation of large language models. *arXiv preprint arXiv:2106.09685*, October 2021.

[8] Renren Jin, Jiangcun Du, Wuwei Huang, Wei Liu, Jian Luan, Bin Wang, and Deyi Xiong. A comprehensive evaluation of quantization strategies for large language models, 2024.

[9] Mike Lewis, Yinhan Liu, Naman Goyal, Marjan Ghazvininejad, Abdelrahman Mohamed, Omer Levy, Ves Stoyanov, and Luke Zettlemoyer. Bart: Denoising sequence-to-sequence pre-training for natural language generation, translation, and comprehension. *arXiv preprint arXiv:1910.13461*, October 2019.

[10] Mingjie Liu, Teodor-Dumitru Ene, Robert Kirby, Chris Cheng, Nathaniel Pinckney, Rongjian Liang, Jonah Alben, Himyanshu Anand, Sanmitra Banerjee, Ismet Bayraktaroglu, Bonita Bhaskaran, Bryan Catanzaro, Arjun Chaudhuri, Sharon Clay, Bill Dally, Laura Dang, Parikshit Deshpande, Siddhanth Dhodhi, Sameer Halepete, Eric Hill, Jiashang Hu, Sumit Jain, Ankit Jindal, Brucek Khailany, George Kokai, Kishor Kunal, Xiaowei Li, Charley Lind, Hao Liu, Stuart Oberman, Sujeet Omar, Ghasem Pasandi, Sreedhar Pratty, Jonathan Raiman, Ambar Sarkar, Zhengjiang Shao, Hanfei Sun, Pratik P Suthar, Varun Tej, Walker Turner, Kaizhe Xu, and Haoxing Ren. Chipnemo: Domain-adapted llms for chip design, 2024.

[11] Renqian Luo, Liai Sun, Yingce Xia, Tao Qin, Sheng Zhang, Hoifung Poon, and Tie-Yan Liu. Biogpt: Generative pre-trained transformer for biomedical text generation and mining. *Briefings in Bioinformatics*, 23(6):bbac409, November 2022.

[12] OpenAI. Gpt-4 technical report. March 2024.

[13] Yongliang Shen, Kaitao Song, Xu Tan, Dongsheng Li, Weiming Lu, and Yueting Zhuang. Hugginggpt: Solving ai tasks with chatgpt and its friends in hugging face. *arXiv preprint arXiv:2303.17580*, December 2023.

[14] Thomas Wolf, Lysandre Debut, Victor Sanh, Julien Chaumond, Clement Delangue, Anthony Moi, Pierric Cistac, Tim Rault, Rémi Louf, Morgan Funtowicz, Joe Davison, Sam Shleifer, Patrick von Platen, Clara Ma, Yacine Jernite, Julien Plu, Canwen Xu, Teven Le Scao, Sylvain Gugger, Mariama Drame, Quentin Lhoest, and Alexander M. Rush. Transformers: State-of-the-art natural language processing. *arXiv preprint arXiv:1910.03771*, July 2020.

[15] Mitchell Wortsman, Gabriel Ilharco, Samir Yitzhak Gadre, Rebecca Roelofs, Raphael Gontijo-Lopes, Ari S. Morcos, Hongseok Namkoong, Ali Farhadi, Yair Carmon, Simon Kornblith, and Ludwig Schmidt. Model soups: averaging weights of multiple fine-tuned models improves accuracy without increasing inference time, 2022.

[16] Stanisław Woźniak, Bartłomiej Koptyra, Arkadiusz Janz, Przemysław Kazienko, and Jan Kocoń. Personalized large language models, 2024.

[17] Shijie Wu, Ozan Irsoy, Steven Lu, Vadim Dabravolski, Mark Dredze, Sebastian Gehrmann, Prabhanjan Kambadur, David Rosenberg, and Gideon Mann. Bloomberggpt: A large language model for finance, 2023.

[18] Hongyang Yang, Xiao-Yang Liu, and Christina Dan Wang. Fingpt: Open-source financial large language models, 2023.

[19] Shih yang Liu, Zechun Liu, Xijie Huang, Pingcheng Dong, and Kwang-Ting Cheng. Llm-fp4: 4-bit floating-point quantized transformers. pages 592–605, 2023.